\documentclass{article}

\usepackage[utf8]{inputenc} % allow utf-8 input
\usepackage[T1]{fontenc}    % use 8-bit T1 fonts
\usepackage{hyperref}       % hyperlinks
\usepackage{url}            % simple URL typesetting
\usepackage{booktabs}       % professional-quality tables
\usepackage{amsfonts}       % blackboard math symbols
\usepackage{nicefrac}       % compact symbols for 1/2, etc.
\usepackage{graphicx}

\newcommand{\hlf}{\textstyle \frac{1}{2} \displaystyle}
\newcommand{\reals}{{\rm I\kern-.17em R}}
\def\Reals{{\hbox{$\it I\hskip-3.6pt R$}}}

\def\Reals{{\hbox{$\it I\hskip-3.6pt R$}}}

\newcommand{\bfx}{{\bf x}}
\newcommand{\bfz}{{\bf z}}

\newcommand{\ie}{{\em i.e.}}
\newcommand{\eg}{{\em e.g.}}

\newcommand{\calL}{{\cal L}}

\newcommand{\calN}{{\cal N}}

\newcommand{\keiko}{\stackrel{\triangle}{=}}

\newcommand{\implies}{{\Rightarrow}}

\begin{document}
\title{Improvements to Supervised EM Learning of Shared Kernel Models by Feature Space Partitioning}
\author{Graham W. Pulford (SMIEEE), BandGapAI, France}

\date{21 March 2022}

\maketitle

\begin{abstract}
Expectation maximisation (EM) is usually thought of as an unsupervised learning method for estimating the parameters of a mixture distribution, however it can also be used for supervised learning when class labels are available. As such, EM has been applied to train neural nets including the probabilistic radial basis function (PRBF) network or shared kernel (SK) model. This paper addresses two major shortcomings of previous work in this area: the lack of rigour in the derivation of the EM training algorithm; and the computational complexity of the technique, which has limited it to low dimensional data sets. We first present a detailed derivation of EM for the Gaussian shared kernel model PRBF classifier, making use of data association theory to obtain the complete data likelihood, Baum's auxiliary function (the E-step) and its subsequent maximisation (M-step). To reduce complexity of the resulting SKEM algorithm, we partition the feature space into $R$ non-overlapping subsets of variables. The resulting product decomposition of the joint data likelihood, which is exact when the feature partitions are independent, allows the SKEM to be implemented in parallel and at $R^2$ times lower complexity. The operation of the partitioned SKEM algorithm is demonstrated on the MNIST data set and compared with its non-partitioned counterpart. It eventuates that improved performance at reduced complexity is achievable. Comparisons with standard classification algorithms are provided on a number of other benchmark data sets.
\end{abstract}


\section{Introduction}
The expectation maximisation algorithm is the main method for learning the parameters of a finite mixture distribution in a numerically tractable way. In essence, it introduces a set of auxiliary variables called ``hidden'' or ``latent'' variables, via which the joint PDF is most easily expressed. The hidden variables provide the categorical information that assigns to each data sample a component in the mixture model \cite{Titterington}. Since these variables are unknown, an iterative approach is adopted wherein the hidden variables are ``averaged out'' of the joint density at each iteration. The parameters are re-estimated by optimising the resulting averaged density and the process is repeated until convergence.  The EM algorithm automatically clusters the data into a set number of component densities whose weights $\pi_k$ are the membership probabilities for the categories represented by the component densities.

Although not widely recognised, the EM algorithm can be viewed as an iterative {\em data association technique}, with the hidden variables playing the role of {\em data association hypotheses}. The degree of membership of each data sample to the mixture components is quantified by a set of {\em posterior} weights $w_{nk}$ that play the role of {\em association probabilities}, as they are known in the  multiple target tracking (MTT) literature. Thus there is a link between hidden variables in the EM algorithm and target tracking methods like joint probabilistic data association (JPDA) \cite{Fortmann2} and probabilistic multihypothesis tracking (PMHT) \cite{Streit7} in terms of their discrete probabilistic structure.

The main concern of this paper is parameter estimation for {\em shared kernel} models (SKMs) from labelled data, with the resulting supervised learning method being called the shared kernel EM (SKEM) algorithm. This type of model arises when we allow each class-conditioned density to originate from a family of PDFs in which the kernels, or component densities, are shared between the classes $c$, with different classes having different weights. We can represent this as
\begin{equation}\label{pxcj}
{\rm p}(x|c=j,\Theta)=\sum_{k=1}^{K}\pi_{kj}\,{\rm p}(x;\theta_k)
\end{equation}
where $\Theta$ is the set of all parameters $\{\pi_{kj},\theta_k\}$, $\pi_{kj}$ is the prior probability that the data sample $x$ belongs to class $j$ and is associated with component $k$. This flexibile representation allows the PDF of each class to represented as a mixture density, rather than just a component of a mixture density. SKMs are also called probabilistic RBFs. If the kernels are Gaussian, then ${\rm p}(x;\theta_k)=\calN\{x;\mu_k,P_k\}$ and each class-conditioned density is a Gaussian mixture. As opposed to more general probabilistic neural nets (PNNs) \cite{Streit94}, all class-conditioned densities use the same set of kernels, or, in the Gaussian case, all have the same mean vectors $\mu_k$ and covariance matrices $P_k$.

Surveying the literature, we find a number of learning schemes based on the EM algorithm that are related but not directly applicable to the SKM problem. Numerous authors constrain either the structure of the mixture covariance matrices or share the covariance matrix between different model components in a standard EM (e.g., \cite{Dharanipragada,Scrucca2016}). The expectation conditional maximisation (ECM) algorithm of \cite{Ng} applies a factorisation over the hidden variables to simplify the M-step for the hierarchical mixtures of experts network \cite{Jordan94}. Gu \cite{Gu} presents an unsupervised EM algorithm that is similar to ours for a multi-sensor estimation problem. Cheng et al. \cite{Cheng} present a self-organising EM algorithm based on the classification EM (CEM) algorithm of \cite{Celeux}, which assigns each data sample to a mixture component via a MAP rule.

An early form SKM was presented in \cite{Luttrell}, where it is referred to as a ``multiple overlapping mixture distribution.'' The idea was formalised in \cite{Jarrad}, where it was referred to as a ``shared mixture distribution.'' Other discriminative classifiers based on Gaussian mixture models (GMMs) that are relevant in the present context include mixture discriminant analysis (MDA) \cite{Hastie1996} and discriminative GMM (DGMM) \cite{Klautau2003}. MDA assumes a model of the form (\ref{pxcj}) with Gaussian kernels, but with identical covariance matrices for all components $P_k=P,~\forall k$, while DGMMs allow different but diagonal covariance matrices. A further technique, referred to as MClustDA in \cite{Fraley2002,Scrucca2014}, reduces a mixture model obtained from MDA algorithm by varying both the number of components in each class and the form of the covariance matrix in a subsequent processing stage. The actual terminology of ``shared kernel model'' is due to Titsias and Likas \cite{Titsias}, who investigated links to feedforward neural network structures employing radial basis functions (RBFs). This method was subsequently renamed ``class conditional mixture density with constrained component sharing'' \cite{Titsias2003}.

The E-step of the SKEM algorithm requires explicit enumeration of {\em all sequences} of data association hypotheses. This is the approach advocated in the original prescription for EM \cite{Dempster}, which bears quoting: ``The E-step is completed by accumulating over all patterns of missing data.'' With only a verbal description of the E-step for finite mixtures appearing in \cite{Dempster}, treatments of the EM algorithm have tended to ``short-circuit'' the derivation of the Q-function, with traditional treatments relying on  binary 0/1 vectors for the hidden variables. The highly cited monograph \cite{Titterington} completely omits a derivation of the E-step in its presentation on mixture parameter estimation. Subsequent derivations in notable works such as \cite{Jordan94} and \cite{Bishop2009} also utilise binary 0/1 vectors and replace the expectation over the sequence of hidden variables by single expectations over the individual variables. More recently, the multinomial distribution has been invoked to describe the role of the hidden variables \cite{McLachlan2}.

From a data association perspective, it is much more natural to use scalar $K$-valued (categorical) hidden variables. The E-step computation is more direct and remains valid even when there are dependencies. (Such cases were noted in Dempster et al.'s 1977 paper.)
E-steps where sequential expectations cannot be avoided arise frequently in the area of MTT data association. For instance, the EM algorithm in \cite{Pulford2002} calls for the expectation over a discrete-time Gauss Markov process, whereas \cite{Alameda-Pineda} considers a variational EM for a hybrid dynamical system where the latent variable is a finite Markov chain. In neural network training, data association theory has hardly been used in the context of the EM algorithm. In the preceding papers \cite{Jarrad,Titsias,Titsias2003}, the derivation of the E-step was glossed over, resulting in expressions for the complete data likelihood and Baum's $Q$ function that are difficult to interpret. An important exception is Streit \& Luginbuhl \cite{Streit94}, where the E-step was computed for PNN training using a simplified ordering. A central aim of this paper is to fill the gaps in the theory by providing theoretically rigorous, but accessible, derivations of (i) the complete data likelihood; (ii) Baum's auxiliary function (the E-step) and (iii) the subsequent maximisation (M-step) in the case of Gaussian SKMs. 

In section \ref{stdEM} we review some relevant aspects of the standard EM algorithm, covering both the binary 0/1 vector latent variable treatment and the use of scalar categorical latent variables, a point we examine further in Appendix A. This is followed in section \ref{skemdetails} by the derivation of the SKEM algorithm. We introduce class-conditioned set notation that correctly accounts for arbitrary class-indexed partitions of the data set. The proof also includes an inductive step that has been left out of previous treatments. We also point out connections between \cite{Titsias} and the earlier work of \cite{Streit94} that were not recognised at that time. Unlike the detailed derivation of the standard EM algorithm for mixture densities \cite{Bilmes}, we avoid the introduction of extra terms using delta functions.

For training PNNs and PRBFs, the EM algorithm has been shown to have distinct advantages over conventional gradient-based training (back-propagation) \cite{Tian1999,Mak2000}, namely: faster convergence, lower computational overhead and monotonically increasing likelihood. Supervised EM classifiers have been incorporated into commercial products, such as medical scanning and screening devices \cite{Menzies2005}. Yet, there have been no significant applications of EM-based training of PNNs or PRBFs since the early 2000s. While this is partly due to the surge in popularity of convolutional neural networks (CNNs) pioneered by LeCun and others, the advantages attributed to EM algorithm training have only been realised on data sets with relatively low dimensionality, e.g., up to 10-D for \cite{Tian1999}. For PRBFs, in \cite{Titsias}, EM simulations are reported for 5-D and 8-D data, as well as the 33-D UCI ionosphere data set, which we revisit in section \ref{fpb}. For higher dimensions the Gaussian evaluations can cause numerical underflow problems even if implemented with the QR algorithm as suggested in \cite{Streit94}, while complexity scales as the cube of the data dimension. To date, this has limited the application of supervised EM algorithms for SKMs and in MDA to low-dimensional data sets.

In section \ref{skemc} we define a framework for applying the SKEM algorithm on higher dimensional feature data than previously attempted. The joint class-conditioned PDF of the data is assumed to factorise according to a given partitioning of the feature space; a lower dimensional EM algorithm being applied in parallel to each partition. This approach is inspired by the mean field theory (MFT) method, which breaks a multi-dimensional optimisation into smaller subproblems by imposing a factorisation on the joint density function via the so-called product density transform \cite{Ormerod}. We demonstrate this ``partitioned SKEM'' (PSKEM) approach on classification problems in section \ref{NS} with feature space dimensions of up to 150, which can be implemented as 10 parallel 15-D EM algorithms rather than a single SKEM with dimension 150. Our results point to some useful practical properties of the PSKEM algorithm, such as numerical stability, robustness to initialisation settings and partitioning arrangement. Most importantly, {\em improved performance} at {\em reduced complexity} is achievable: that is, the accuracy of the PSKEM is generally higher than that of the non-partitioned SKEM in addition to the significant speed up afforded by partitioning. The degree to which mixture components are shared and the relationship on classifier accuracy are also briefly examined. We provide in section \ref{fpb} comparative performance testing on some popular benchmark data sets. Section \ref{conc} concludes with some directions for future research.

\section{Shared Kernel EM Algorithm}\label{skem}
\subsubsection*{Notation}
The treatment of the SKEM algorithm uses the following notation.
\begin{itemize}
\item Incomplete data $X=\{x_1,\ldots,x_N\},~x_i\in\Reals^M$ is a data point or feature vector.
\item Class labels $C=\{c_1,\ldots,c_N\}$, with each element $c_i\in\{1,\ldots,L\}$, where $L$ is the number of classes.
\item Hidden or missing data $Z=\{z_1,\ldots,z_N\}$, also called latent variables. $z_n=k$ indicates that $x_n$ is associated with mixture component $k$.
\item Complete data $Y=(X,Z)$, the union of the incomplete and hidden data.
\item Parameters $\theta_k$ for component $k$ in a SKM.
\item Full set of parameters $\Theta$, including mixture weights.
\item Estimated parameters $\Theta_p$ at pass $p$ of the algorithm.
\end{itemize}
In the case of a $K$-component multivariate Gaussian SKM, the full parameter set $\Theta$ comprises the matrix of class-conditioned weights $\{\pi_{kj}\}$, $k=1,\ldots,K$, $j=1,\ldots,L$, mean vectors $\{\mu_k\}$, where $\mu_k\in\Reals^M$, and covariance matrices $\{P_k\}$, where $P_k\in\Reals^{M\times M}$. The shared kernel parameters for mixture component $k$ are $\theta_k=\{\mu_k,P_k\}$.

\subsection{Standard EM Algorithm}\label{stdEM}
A brief review of the standard EM algorithm will be useful to highlight the differences between the latter and the SKEM algorithm, as well as the two main techniques used in EM-type derivations.
The EM algorithm seeks the parameter estimates $\widehat{\Theta}$ that maximise the joint data likelihood:
\begin{equation}\label{pxtheta}
{\rm p}(X|\Theta)=\prod_{i=1}^N {\rm p}(x_i;\Theta)=\prod_{i=1}^N
\sum_{k=1}^K \pi_k \calN\{x_i; \mu_k,P_k\}
\end{equation}
in which $\calN\{x;\mu_k,P_k\}$ is a $M$-D Gaussian PDF with mean $\mu_k$ and covariance $P_k$. Since direct maximisation is usually intractable, EM works instead with the complete data likelihood ${\rm p}(Y|\Theta)$. Conditioning on the latent variables leads to:
\begin{equation}\label{pycat}
{\rm p}(Y|\Theta)=\prod_{i=1}^N{\rm p}(x_i|z_i,\Theta)
\Pr(z_i|\Theta)=\prod_{i=1}^N \pi_{z_i}\calN\{x_i;\mu_{z_i},P_{z_i}\}
\end{equation}
This at once makes clear the rationale for forming the complete data likelihood, revealing the role of the $z_i$ as {\em data association} variables. Taking the logarithm of this results in a more manageable expression:
\[
\log\,{\rm p}(Y|\Theta)=\sum_{i=1}^{N}\log\,\pi_{z_i}+\sum_{i=1}^{N}\log\,\calN\{x_i;\mu_{z_i},P_{z_i}\}
\]
The EM algorithm then evaluates Baum's auxiliary function:
\begin{eqnarray}
Q(\Theta,\Theta_0)&=&{\rm E}_Z[\log {\rm p}(X,Z|\Theta) | X,\Theta_0]\label{baum2a}\\
&=&\sum_{z_1=1}^{K}\cdots\sum_{z_N=1}^{K}\log{\rm p}(X,Z|\Theta)\Pr(Z|X,\Theta_0)\nonumber
\end{eqnarray}
in the E-step, which is followed by its maximisation in the M-step.

By contrast, the traditional approach \cite{Titterington} defines indicator vectors $\bfz_i$, $i=1,\ldots,N$ of size $K$ with binary elements $z_{ij}\in\{0,1\}$, in which only element $j$ equals 1. In this scheme, the complete data likelihood is written as:
\begin{equation}\label{titlik}
{\rm p}(Y|\Theta)=\prod_{i=1}^{N}\prod_{j=1}^{K}(\pi_j\,{\rm p}(x_i;\theta_j))^{z_{ij}}
\end{equation}
The economy of notation obtained by raising a PDF to a power only makes sense if $z_{ij}$ is binary. (The inclusion of vector variables $\bfz_i$ is in fact redundant and leads to dependencies between the components---a point we explore further in Appendix A.) The $Q$ function is derived by noting that the expectation of a 0/1 random variable is equal to the probability that it equals 1. The EM algorithm is then obtained by maximising the $Q$ function, leading to the iterative scheme below, where $p$ is the pass:
\begin{eqnarray}
w^{(p)}_{nk} &=& \frac{\pi^{(p-1)}_k {\calN}\{x_n;\mu^{(p-1)}_k,P^{(p-1)}_k\}}{\sum_{k=1}^{K}\pi^{(p-1)}_k{\calN}\{x_n;\mu^{(p-1)}_k,P^{(p-1)}_k\}} \label{emw}\\
\pi^{(p)}_k &=& \frac{1}{N}\sum_{n=1}^{N}w^{(p)}_{nk} \label{empi}\\
\mu^{(p)}_k &=& \frac{\sum_{n=1}^{N}w^{(p)}_{nk}x_n}{\sum_{n=1}^{N}w^{(p)}_{nk}} \label{emmu}\\
P^{(p)}_k &=& \frac{\sum_{n=1}^{N}w^{(p)}_{nk}(x_n-\mu^{(p)}_k)(x_n-\mu^{(p)}_k)^T}{\sum_{n=1}^{N}w^{(p)}_{nk}} \label{emP}
\end{eqnarray}
A proof using categorical variables may be found in \cite{Pulford2020b}.

\subsection{SKEM Algorithm Details}\label{skemdetails}
The class-conditioned density ${\rm p}(x|c=j,\Theta)$ for the SKEM algorithm, for some $x\in\Reals^M$, is assumed to take the form  (\ref{pxcj}) corresponding to a Gaussian mixture with parameters $\Theta$. We can interpret the weight $\pi_{kj}$ as the probability that a data point is associated with component $k$ of the mixture density conditioned on the class $j$. We will also write the association hypothesis in the form $z_n=k$ for some data point $x_n$.

The objective is to determine the maximum likelihood estimate of the SK density parameters and weights such that:
\begin{equation}\label{em2}
    \widehat{\Theta}=\arg\max_{\Theta} {\rm p}(X|C,\Theta)
\end{equation}
which differs from the conventional EM criterion (\ref{pxtheta}) by the presence of the class labels in the conditioning on the right side. The latter makes the SKEM problem a supervised learning problem. The key to solving the optimisation problem is to introduce so-called hidden variables $Z$ that cover all possible association hypotheses between the data and the mixture components, given the class to which the particular data point pertains. The association mechanism is analogous to that used in MTT, where there is a need to associate measurements with target objects. Given an initial parameter estimate $\Theta_0$, Baum's auxiliary function is defined as in (\ref{baum2a}) by:
\begin{eqnarray}\label{baumc}
    Q(\Theta,\Theta_0)&=&{\rm E}_Z[\log {\rm p}(X,Z|C,\Theta) | X,\Theta_0]\label{baum2d}\\
    &=&\sum_{z_1=1}^{K}\cdots\sum_{z_N=1}^{K}\log{\rm p}(X,Z|C,\Theta)\Pr(Z|X,\Theta_0)\nonumber
\end{eqnarray}
The logarithm of the complete data likelihood decomposes via the {\em iid} assumption on the data into a sum of factors in the E-step. This is maximised in the M-step to yield parameter updating equations. The calculations are more complicated than the standard EM derivation due to the presence of class conditioning, which requires more specialised ``accounting.'' We first state the result below and then deal with the proof.

\subsubsection*{Statement of the SKEM algorithm for Gaussian mixtures}
The class vector $C$ partitions the data set into non-overlapping subsets $X_i$, containing the feature vectors for class $i$, with cardinality $l(i)=|X_i|$ defined by
\begin{equation}\label{skemXi}
X_i=\{x_j,~j\in\Gamma_i\},~i=1,\ldots,L
\end{equation}
where $\Gamma_i$ the index set for class $i$, that is, the set of indexes from $\{1,\ldots,N\}$ corresponding to all the class $i$ feature vectors. Note that the sum of the $l(i)$ is the sum of the numbers of feature vectors in each of the $L$ classes, namely $l(1)+\cdots+l(L)=N$. The missing data $Z$ are partitioned as:
\begin{equation}\label{skemZi}
Z_i=\{z_j,~j\in\Gamma_i\},~i=1,\ldots,L
\end{equation}
It is convenient to write the elements of the index set $\Gamma_i$ in the following way
\begin{equation}\label{skemGamma}
\Gamma_i=\{n_{ij}\},~i=1,\ldots,L,~j=1,\ldots,l(i)
\end{equation}
Clearly, the $\Gamma_i$ are disjoint and every feature vector belongs to one and only one class, so we have $\bigcup_{i=1}^{L}\Gamma_i=\{1,\ldots,N\}.$
Let $\Theta_0=\{\pi^{(0)}_{kj}, \mu^{(0)}_k,P^{(0)}_k\}_{k=1\!,\,j=1}^{K,~~L}$, be an initial estimate of the SKM parameters. The SKEM algorithm comprises the following iterative steps, performed in order, starting with $p=1$ and incrementing $p$ till convergence. The presence of both indexes $n_{ij}$ and $n$ in the equations is deliberate.
\begin{eqnarray}
w^{(p)}_{n_{ij}k}&=&\frac{\pi^{(p-1)}_{ki} \calN\{x_{n_{ij}};\,\mu^{(p-1)}_k,P^{(p-1)}_k\}}{\sum_{k=1}^{K}\pi^{(p-1)}_{ki}\calN\{x_{n_{ij}};\,\mu^{(p-1)}_k,P^{(p-1)}_k\}} \label{skemw}\\
\pi^{(p)}_{ki} &=& \frac{1}{l(i)}\sum_{n\in\Gamma_i}w^{(p)}_{nk} \label{skempi}\\
\mu^{(p)}_k &=& \frac{\sum_{n=1}^{N}w^{(p)}_{nk}x_n}{\sum_{n=1}^{N}w^{(p)}_{nk}} \label{skemmu}\\
P^{(p)}_k &=& \frac{\sum_{n=1}^{N}w^{(p)}_{nk}(x_n-\mu^{(p)}_k)(x_n-\mu^{(p)}_k)^T}{\sum_{n=1}^{N}w^{(p)}_{nk}} \label{skemP}
\end{eqnarray}
At all iterations, the class-conditioned mixture weights $\pi_{ki}$ and posterior weights (association probabilities) $w_{nk}$ satisfy the normalisation conditions
$\sum_{k=1}^{K}\pi_{ki}=1$, $\sum_{k=1}^{K}w_{nk}=1$. The resemblance of the SKEM algorithm to standard unsupervised EM (\ref{emw})--(\ref{emP}) is deceptive since the equations have the same form. Closer examination reveals that for SKEM, the $\pi_{ki}$ have two indexes rather than one, and the $w_{nk}$ have class-conditioned indexing. The connection to standard EM is clearer here than in \cite{Titsias}, which has double summations on the $\mu_k$ and $P_k$ update equations.

\subsection*{Proof Outline}
The proof is split into two parts. The first part deals with the E-step, the second with the M-step.

\subsubsection*{Evaluation of Baum's Auxiliary Function for the SKEM}
We start by expressing the joint likelihood function for the complete data $Y=(X,Z)$ by partitioning the data set according to the class $c_i$. On account of the {\em iid} assumption on the feature vectors, and bearing in mind the definitions in (\ref{skemXi}), (\ref{skemZi}) and (\ref{skemGamma}), we have
\begin{eqnarray}
{\rm p}(X,Z|C,\Theta) &=& \prod_{i=1}^{L}\Pr(Z_i|c_i,\Theta)\,{\rm p}(X_i|Z_i,c_i,\Theta)\nonumber\\
 &=& \prod_{i=1}^{L}\prod_{j=1}^{l(i)}\Pr(z_{n_{ij}}|c_i,\Theta)\,{\rm p}(x_{n_{ij}}|Z_i,c_i,\Theta)\nonumber\\
 &=& \prod_{i=1}^{L}\prod_{j=1}^{l(i)}\Pr(k_{ij}|c_i,\Theta)\,{\rm p}(x_{n_{ij}}|\theta_k)
\end{eqnarray}
where, for subsequent notational convenience, we have defined $k_{ij} = z_{n_{ij}}$, being the component in the mixture assigned to feature vector $x_{n_{ij}}$. Note that the decomposition into products {\em not} involving sums of terms dependent on the parameters is crucial to the simplification of the log likelihood. Taking the logarithm gives $\log {\rm p}(X,Z|C,\Theta)$ as a sum of simple terms:
\begin{equation}\label{logpxzc}
\sum_{i=1}^{L}\sum_{j=1}^{l(i)}\log\left(\Pr(k_{ij}|c_i,\Theta)\,{\rm p}(x_{n_{ij}}|\theta_k)\right)
\end{equation}
At this point we make a number of definitions that will be useful in the sequel. The indexes ranges are $k=1,\ldots,K$; $c=1,\ldots,L$; $j=1,\ldots,l(i)$.
\begin{eqnarray}
\pi_{kc} \!\!\!&=&\!\!\! \Pr(z=k|c,\Theta) \label{pikc}\\
g_{ij}\!\!\!&\keiko&\!\!\! g(k_{ij},x_{n_{ij}})= \log\Pr(k_{ij}|c_i,\Theta)+\log{\rm p}(x_{n_{ij}}|\theta_k)\nonumber\\
h_{ij}\!\!\!&\keiko&\!\!\! h(k_{ij}|x_{n_{ij}})=\Pr(k_{ij}|x_{n_{ij}},c_i,\Theta_0) \label{hij}
\end{eqnarray}
Since both $\pi_{kc}$ and $h_{ij}$ are probability mass functions (PMFs), they satisfy the normalisation conditions
\begin{equation}\label{norm}
\sum_{k=1}^{K}\pi_{kc}=1,~\sum_{k_{ij}=1}^{K}h(k_{ij}|x_{n_{ij}})=1
\end{equation}
The reason for introducing the $h_{ij}$ functions is to express the joint PMF of the latent variables given the data and initial parameter estimates, {\em viz.}:
\begin{eqnarray}
\Pr(Z|X,\Theta_0) &=& \prod_{i=1}^{L}\Pr(Z_i|X_i,\Theta_0)\label{pzxtheta0}\\
&=& \prod_{i=1}^{L}\prod_{j=1}^{l(i)}\Pr(z_{n_{ij}}|x_{n_{ij}},c_i,\Theta_0)\nonumber\\
&=&\prod_{i=1}^{L}\prod_{j=1}^{l(i)}h(k_{ij}|x_{n_{ij}})\keiko\prod_{i=1}^{L}\prod_{j=1}^{l(i)}h_{ij}\nonumber
\end{eqnarray}
The construction of Baum's auxiliary function requires the expectation over the {\em sequence} of latent variables to be calculated. Since the latter are discrete, an $N$-fold sum over the components $z_i$ of $Z$ is involved. Each $z_i$ can take any of $K$ values; thus there are $K^N$ terms in the expectation in equation (\ref{baum2d}). The class conditioning allows this $N$-fold sum to be rearranged as $L$ groups of sums each having $l(i)$ individual sums. This corresponds to the factorisation $K^{l(1)}\times\cdots K^{l(L)}=K^N$. With the preceding notational definitions and equations (\ref{logpxzc}) and (\ref{pzxtheta0}), it is now possible to write $Q(\Theta,\Theta_0)$ from (\ref{baum2d}) as:
\begin{eqnarray}
Q(\Theta,\Theta_0)\!\!\!\!\!&=&\!\!\!\!\!
\underbrace{\left(\sum_{k_{11}=1}^{K}\cdots\sum_{k_{1l(1)}=1}^{K}\right)}_{l(1)\mbox{ sums}}
\cdots
\underbrace{\left(\sum_{k_{L1}=1}^{K}\cdots\sum_{k_{Ll(L)}=1}^{K}\right)}_{l(L)\mbox{ sums}} (\star) \nonumber\\
(\star) \!\!\!&\keiko&\!\!\! \sum_{i=1}^{L}\sum_{j=1}^{l(i)}g_{ij}
\prod_{m=1}^{L}\prod_{r=1}^{l(m)}h_{mr}\label{Lsums}
\end{eqnarray}
Rather than deal with the general case, we present the simplification of the result for the $L=2$ class case. The general case follows by induction. Setting $L=2$ in (\ref{Lsums}), we get
\[
Q(\Theta,\Theta_0)=
\underbrace{\left(\sum_{k_{11}=1}^{K}\cdots\sum_{k_{1l(1)}=1}^{K}\right)}_{l(1)\mbox{ sums}}
\underbrace{\left(\sum_{k_{21}=1}^{K}\cdots\sum_{k_{2l(2)}=1}^{K}\right)}_{l(2)\mbox{ sums}}(\dagger)
\]
where
\begin{equation}
(\dagger)\keiko
\sum_{i=1}^{2}\sum_{j=1}^{l(i)}g_{ij}
\prod_{m=1}^{2}\prod_{r=1}^{l(m)}h_{mr}\label{Lsums2}
\end{equation}
Now define
\[
T_{ij}\!=\!\!\underbrace{\left(\sum_{k_{11}=1}^{K}\!\cdots\!\sum_{k_{1l(1)}=1}^{K}\right)}_{l(1)\mbox{ sums}}
\underbrace{\left(\sum_{k_{21}=1}^{K}\!\cdots\!\sum_{k_{2l(2)}=1}^{K}\right)}_{l(2)\mbox{ sums}}
g_{ij}\!\prod_{m=1}^{2}\prod_{r=1}^{l(m)}h_{mr}
\]
so that
\[
Q(\Theta,\Theta_0)=\sum_{i=1}^{2}\sum_{j=1}^{l(i)}T_{ij}
\]
Re-express the first term on the right hand side as
\[
T_{11}\!=\sum_{k_{11}}g_{11}h_{11}\!
\underbrace{\left(\sum_{k_{12}}\!\cdots\!\sum_{k_{1l(1)}}\right)}_{l(1)-1\mbox{ sums}}
\underbrace{\left(\sum_{k_{21}}\!\cdots\!\sum_{k_{2l(2)}}\right)}_{l(2)\mbox{ sums}}
\!\prod_{(m,r)\neq(1,1)}\!\!\!h_{mr}
\]
in which for notational convenience we have omitted the limits in the sums. We treat the right hand part of $T_{11}$ (after the group of $l(1)-1$ sums involving $k_{1j}$) in a similar manner to the proof of the standard EM algorithm E-step \cite{Pulford2020b}:
\begin{eqnarray*}
 && \underbrace{\left(\sum_{k_{21}}\cdots\sum_{k_{2l(2)-1}}\right)}_{l(2)-1\mbox{ sums}}
\left(\sum_{k_{2l(2)}}h_{2l(2)}\right)
\prod_{{(m,r)\neq(1,1)}\atop{(m,r)\neq(2,l(2))}}h_{mr} \\
 &=& \underbrace{\left(\sum_{k_{21}}\cdots\sum_{k_{2l(2)-1}}\right)}_{l(2)-1\mbox{ sums}}
\prod_{{(m,r)\neq(1,1)}\atop{(m,r)\neq(2,l(2))}}h_{mr}
\end{eqnarray*}
where we used the normalisation condition (\ref{norm}). Proceeding by induction on both indexes $k_{1j}$ and $k_{2j}$, we obtain (somewhat miraculously):
\[
\underbrace{\left(\sum_{k_{12}}\cdots\sum_{k_{1l(1)}}\right)}_{l(1)-1\mbox{ sums}}
\underbrace{\left(\sum_{k_{21}}\cdots\sum_{k_{2l(2)}}\right)}_{l(2)\mbox{ sums}}
\prod_{(m,r)\neq(1,1)}h_{mr}=1
\]
which means that
\[
T_{11}=\sum_{k_{11}=1}^{K}g_{11}h_{11}=\sum_{k_{11}=1}^{K}g(k_{11},x_{n_{11}})\,h(k_{11}|x_{n_{11}})
\]
It follows immediately that
\[
T_{ij}=\sum_{k_{ij}=1}^{K}g_{ij}h_{ij}=\sum_{k_{ij}=1}^{K}g(k_{ij},x_{n_{ij}})\,h(k_{ij}|x_{n_{ij}})
\]
whence, in the $L=2$ case
\[
Q(\Theta,\Theta_0)=\sum_{i=1}^{2}\sum_{j=1}^{l(i)}\sum_{k_{ij}=1}^{K}g(k_{ij},x_{n_{ij}})\,h(k_{ij}|x_{n_{ij}})
\]
and, in the general $L$-class case:
\begin{equation}\label{baum2e}
Q(\Theta,\Theta_0)=\sum_{i=1}^{L}\sum_{j=1}^{l(i)}\sum_{k_{ij}=1}^{K}g(k_{ij},x_{n_{ij}})\,h(k_{ij}|x_{n_{ij}})
\end{equation}
In light of definitions (\ref{pikc})--(\ref{hij}), a few tweaks can be made to tidy up this last expression for the auxiliary function. Concentrating on $h_{ij}$, we apply Bayes' rule to obtain
\[
\Pr(k_{ij}|x_{n_{ij}},c_i,\Theta_0)=\frac{\Pr(k_{ij}|c_i,\Theta_0)\,{\rm p}(x_{n_{ij}}|k_{ij},\Theta_0)}
{\sum_{k_{ij}=1}^{K}\Pr(k_{ij}|c_i,\Theta_0)\,{\rm p}(x_{n_{ij}}|k_{ij},\Theta_0)}
\]
where, in the Gaussian mixture case,
\[
{\rm p}(x_{n_{ij}}|k_{ij},\Theta_0)=
\calN\{x_{n_{ij}};\,\mu^{0}_{k_{ij}},P^{0}_{k_{ij}}\}
\]
Upon replacing the index $k_{ij}$ by $k$, reintroducing $\pi_{kc}$ and redefining $h(k|x_{n})$
as $w_{nk}$, we obtain
\begin{equation}\label{wnijk}
w_{n_{ij}k}=\frac{\pi^{0}_{ki}\,\calN\{x_{n_{ij}};\,\mu^{0}_k,P^{0}_k\}}{\sum_{k=1}^{K}\pi^{0}_{ki}\,\calN\{x_{n_{ij}};\,\mu^{0}_k,P^{0}_k\}}
\end{equation}
which we recognise as equation (\ref{skemw}). For Baum's auxiliary function in (\ref{baum2d}) we finally obtain the simplified expression
\begin{equation}\label{baums}
Q(\Theta,\Theta_0)=\sum_{i=1}^{L}\sum_{j=1}^{l(i)}\sum_{k=1}^{K} w_{n_{ij}k}
\left(\log\pi_{ki}+\log\calN(x_{n_{ij}};\mu_k,P_k)\right)
\end{equation}
It can be appreciated from the preceding arguments that obtaining $Q(\Theta,\Theta_0)$ by sequential summation is nontrivial. (\ref{baums}) should be compared with equation (15) in \cite{Titsias} (whose different notational conventions we modify for compatibility), where the auxiliary function is stated without proof in terms of a quantity $\Pr(k|c_i,x_n)$ that has 3 indexes: $k$ (component), $i$ (class) and $n$ (sample). It can be shown that the algorithm in \cite{Titsias} is in fact the same as the SKEM presented here, but the connection is not obvious due to the differences in indexing. By using a sample index $n$ that depends on both the class $i$ and the index $j$ within the class-conditioned set of $l(i)$ elements, we arrive at a clearer statement of the algorithm that is simpler to understand and more easily implementable.

The EM training algorithm for PNNs in \cite{Streit94}, which computes the expectation over all sequences of latent variables $Z$, contains (\ref{baums}) as a special case due to the 4-layer structure of a PNN, which is capable of modelling a SKM with its first 3 layers. On the other hand, the M-steps for a PNN and for a SKM are different due to the presence of equality constraints on the mixture components in the ``pattern layers.'' Specifically, for equivalence of a SKM with a PNN (in Streit and Luginbuhl's notation), we require the {\em same} number of components in each ``pool,'' {\em different} covariance matrices $\Sigma$ and {\em equal} mean vectors $\mu_{ij}=\mu_j$ $\forall i$. The M-step we derive next accounts for these aspects.

\subsubsection*{Optimisation of Baum's Auxiliary Function for the SKEM}
We now compute the M-step for the SKEM algorithm, having already expressed Baum's auxiliary function $Q(\Theta,\Theta_0)$ in (\ref{baums}). We retain the simpler notation regarding the pass or iteration, with $\Theta$ denoting parameters at the current pass and $\Theta_0$ for parameters from the previous pass (including initialisation). The optimisation problem for the M-step is:
\begin{equation}\label{Mstep}
\max_{\Theta}{Q(\Theta,\Theta_0)} \mbox{ subject to } \sum_{k=1}^{K}\pi_{kc}=1
\end{equation}
The equality-constrained optimisation problem is first expressed as an unconstrained problem by introducing Lagrange multipliers $\lambda=(\lambda_1,\ldots,\lambda_L)^T$, one per constraint on the parameters to be estimated:
\[
\calL(\Theta,\lambda_1,\ldots,\lambda_L;\Theta_0)=
Q(\Theta,\Theta_0)-\sum_{i=1}^{L}\lambda_i\left(\sum_{k=1}^{K}\pi_{ki}-1\right)
\]
First-order necessary conditions for a maximum require $\nabla_{\Theta}\calL={\bf 0}$ and $\nabla_{\lambda}\calL={\bf 0}$. Considering first the class-conditioned weights (which do not depend on the form of the conditional PDFs), we must have
\[
\frac{\partial \calL}{\partial\pi_{ki}}=\frac{1}{\pi_{ki}}\sum_{j=1}^{l(i)}w_{n_{ij}k}-\lambda_i=0
\]
and also
\[
\frac{\partial \calL}{\partial\lambda_i}=0~\implies\sum_{j=1}^{l(i)}\sum_{k=1}^{K} w_{n_{ij}k}=\lambda_i
\]
which, because of the normalisation condition, gives $\lambda_i=l(i)$. It follows that
\[
\pi_{ki}=\frac{1}{l(i)}\sum_{j=1}^{l(i)}w_{n_{ij}k}
\]
Recalling the definition of the index set $\Gamma_i$ from (\ref{skemGamma}), this is more naturally expressed as
\begin{equation}
\pi_{ki}=\frac{1}{l(i)}\sum_{n\in\Gamma_i}w_{nk}
\end{equation}
which we recognise as the update equation for $\pi_{ki}$ in (\ref{skempi}).

We borrow some results from \cite{Petersen} (section 8.4) concerning the derivatives of the logarithm of a multivariate Gaussian PDF with respect to its mean vector and covariance matrix:
\begin{eqnarray*}
\frac{\partial}{\partial\mu}\log\calN\{x;\,\mu,P\} &=& P^{-1}(x-\mu)\\
\frac{\partial}{\partial P}\log\calN\{x;\,\mu,P\} &=& \hlf P^{-1}
(x-\mu)(x-\mu)^TP^{-1}-\hlf P^{-1}
\end{eqnarray*}
With these results in hand, the remaining update equations are easy to obtain. For the mean vectors:
\[
\frac{\partial \calL}{\partial\mu_k}=\sum_{i=1}^{L}\sum_{j=1}^{l(i)} w_{n_{ij}k}P^{-1}_k(x_{n_{ij}}-\mu_k)
={\bf 0}
\]
Now pre-multiply by $P_k$ and solve for $\mu_k$
\[
\mu_k = \frac{\sum_{i=1}^{L}\sum_{j=1}^{l(i)} w_{n_{ij}k}\,x_{n_{ij}} }{\sum_{i=1}^{L}\sum_{j=1}^{l(i)} w_{n_{ij}k}}
\]
Noting that $\sum_{i=1}^{L}\sum_{j=1}^{l(i)}f(n_{ij})=\sum_{n=1}^{N}f(n)$, the latter is the same as (\ref{skemmu}).

For the covariance matrices: $\frac{\partial \calL}{\partial P_k}={\bf 0}~\implies$
\[
\sum_{i=1}^{L}\sum_{j=1}^{l(i)} w_{n_{ij}k}
P^{-1}_k\left[(x_{n_{ij}}-\mu_k)(x_{n_{ij}}-\mu_k)^T P^{-1}_k-I\right]={\bf 0}
\]
Multiply the above by $P_k$ on the left and right to give
\[
\sum_{i=1}^{L}\sum_{j=1}^{l(i)} w_{n_{ij}k}
\left[(x_{n_{ij}}-\mu_k)(x_{n_{ij}}-\mu_k)^T-P_k\right]={\bf 0}
\]
which yields
\[
P_k = \frac{\sum_{i=1}^{L}\sum_{j=1}^{l(i)} w_{n_{ij}k}\,(x_{n_{ij}}-\mu_k)(x_{n_{ij}}-\mu_k)^T }{\sum_{i=1}^{L}\sum_{j=1}^{l(i)} w_{n_{ij}k}}
\]
Rejigging the indexing in the previous equation gives the covariance update in (\ref{skemP}).
For subsequent comparison, we mention that the MDA algorithm differs from the SKEM algorithm through the assumption of identical component covariances. The covariance update for MDA is easily shown to be (for all components $k$ at pass $p$):
\begin{equation}\label{mdacov}
P^{(p)}=\frac{1}{N}\sum_{k=1}^{K}\sum_{n=1}^{N}w^{(p)}_{nk}(x_n-\mu^{(p)}_k)(x_n-\mu^{(p)}_k)^T
\end{equation}

\subsection*{SKEM Algorithm Complexity}
Retaining the notation from section \ref{skem}, the SKEM training algorithm for $M$-dimensional features, $K$ components, $N$ training samples and $N_P$ passes has a constant computational complexity of $N_P K N (2+M+M^2+g(M))$ where $g(M)$ is the complexity of the quadratic form evaluation in the Gaussian PDF, which is an O($M^3$) operation. Thus the overall complexity is O($N_PKNM^3$). The memory requirements are $(2N+M+L)K+(M+1)N$ for $L$ classes.

\subsection{Reduced Complexity SKEM Classifier}\label{skemc}
A MAP classifier can be constructed from the family of class-conditioned shared kernel PDFs learned from the data by the SKEM algorithm. In the case of a uniform prior distribution across the classes $\Pr(c)=1/L$, which we assume here, the MAP criterion reduces to a ML criterion:
\begin{equation}
\hat{c}_{\rm ML}=\arg\max_{c\in\{1,\ldots,L\}}{\rm p}(x|c=j)
\end{equation}
In the case of shared kernel PDFs this gives:
\begin{equation}\label{skcl}
\hat{c}_{\rm ML}={\arg\max}_{j\in\{1,\ldots,L\}}\sum_{k=1}^{K}\pi_{kj}\,\calN\{x;\mu_k,P_k\}
\end{equation}
which we refer to as the shared kernel classifier (SKC).

Learning the parameters of a SKC has complexity O($N_PKNM^3$). For high dimensional data sets, \eg, $M={\rm dim}(x)>100$, the SKEM algorithm is not implementable without specialised hardware and would likely have poor numerical stability. Consider a partition of the feature vector according to $\bfx=[x_1^T,\ldots,x_R^T]^T$, where, without loss of generality, the partitions $\rho$ have equal dimension so $x_r\in\Reals^m$, $r=1,\ldots,R$, where $M=mR$. Suppose further that the $x_r$ are {\em mutually independent} given the class. Defining $q_{rj}(x)={\rm p}(x|c=j,\rho=r)$, the following factorisation is then valid:
\begin{equation}\label{pskemfact}
{\rm p}(\bfx|c=j)=\prod_{r=1}^R q_{rj}(x_r),~j=1,\ldots,L
\end{equation}
Note that mutual independence requires more than a simple division of the feature space into non-overlapping subsets, as might be the case in a mean field theoretic decomposition. The lower dimensional PDFs $q_{rj}(x)$ are different SKMs, each having $K$ components. The product of independent Gaussian PDFs in the variables $\{x_1,\ldots,x_R\}$ can be expressed as a single Gaussian PDF in $\bfx$, however, there are $K^R$ such Gaussian components in the factorised PDF (\ref{pskemfact}) as opposed to $K$ in the original mixture model (\ref{pxcj}). This point is clarified in Appendix B.

The $q_{rj}(x)$ are learned by the SKEM  algorithm on the $N\times m$ partition of training data $X^{(r)}=\{\bfx_{r1},\ldots,\bfx_{rN}\}$, producing the partitioned SKMs $\{\pi_{kj}^{(r)},\mu_k^{(r)},P_k^{(r)}\}$ for $k=1,\ldots,K$, $j=1,\ldots,L$, $r=1,\ldots,R$. Since the data partitions are non-overlapping, the training can be carried out in parallel, leading to a parallel or partitioned SKEM (PSKEM) algorithm. This reduced-complexity PSKEM classifier has class-conditioned log likelihood function (with $\bfx\in\Reals^M$)
\begin{equation}\label{skemp}
\hspace{-3mm}
\log{\rm p}(\bfx|c=j)=\sum_{r=1}^{R}\log\left\{
\sum_{k=1}^{K} \pi_{kj}^{(r)} \calN\{x_r;\mu_k^{(r)},P_k^{(r)}\}
\right\}
\end{equation}
Like the MFT approach to variational optimisation, the partitioned SKEM algorithm just described requires iteration. However, unlike MFT, each partitioned SKEM can be iterated independently until convergence. There is no reason that the same number of passes should be required for convergence of each set of SK parameters. Notwithstanding, for ease of implementation, the version given in this paper performs one pass at a time on each partition, constructs the log likelihoods in (\ref{skemp}) and validates the model on a test data set before performing the next pass. This approach, which is also used in deep learning of convnets, helps to assess algorithm performance ``on the fly'' and to avoid overfitting, \ie, too close an approximation to a given training data set.

The computational complexity of the partitioned SKEM training algorithm is O($RN_PKNm^3$), which is $R^2$ times less than the original $M$-D SKEM. In section \ref{NS}, we give an example where $M=36$ and $R=1$ to $R=12$. In this case, the partitioned SKEM has 1-2 orders of magnitude lower compelxity than its 36-D counterpart, while outperforming it in classification accuracy on the same feature data.

\section{Implementation \& Numerical Simulations}\label{NS}
\subsection{Performance Benchmark: MNIST}
The SKEM algorithm was applied to the handwritten digit recognition problem using the well known MNIST database \cite{LeCun2}. This data set contains 60,000 training and 10,000 test greyscale images all of size 28 x 28 pixels. The following pre-processing was performed for each image: (i) de-skewing via affine transformation based on image moments up to order two as in \cite{LeCun}; (ii) range scaling to [0,1]; (iii) normalisation by subtraction of the mean and division by the standard deviation. Data reduction was achieved via singular value decomposition (SVD) of the sample covariance of the training data as in \cite{Er}. PCA features were obtained by subtraction of the mean (of the training data) followed by projection of the vectorised data onto the $r=150$ most significant eigenvectors from the SVD. Although the partitioned features are not independent, they are partially decorrelated and serve to demonstrate the application of the SKEM classifier. The partitions were obtained in a sequential manner from the PCA features.

The partitioned SKEM algorithm was coded in Matlab$^{\rm TM}$. The detailed implementation is covered  in \cite{Pulford2020a}. A random subset of 30,000 of training images was used. On each $m$-dimensional feature partition (or block), initial mixture means $\mu_k$ were {\em iid} uniform on [$-2$, 2]; initial mixture covariance matrices $P_k$ were taken as $\sigma^2 I_m$ where $\sigma=2$; initial mixing probabilities $\pi_{kj}$ were uniform for each of the 10 classes. After one pass on each feature partition, the PSKEM classifier (\ref{skemp}) was evaluated on the test data and the confusion matrix and mean accuracy calculated. PSKEM performance generally peaked around 10 - 25 such passes, indicating convergence. The training was run for $N_P=30$ passes and the best mean accuracy noted. Experiments were carried out for $R$ partitions of $m$-dimensions with $mR$ in the range 30 - 150 (being the number of extracted PCA features). The number of mixture components $K$ in each SKM (assumed to be the same on each partition) was varied in the range 4 to 60. 

The peak accuracy obtained was 97.48\% using $R=3$ partitions of dimension $m=13$ and $K=60$ components, achieved at 24 passes. The choice of 39 PCA features was motivated by the benchmark polynomial support vector machine (SVM) result from \cite{LeCun}, which used 40 PCA features and achieved 96.7\% accuracy on MNIST. While the accuracy obtained is less than that which can be achieved with deep convolutional neural networks, it benchmarks supervised PSKEM learning in the list of techniques that have been applied to MNIST.

\subsection{Further Performance Benchmarks}\label{fpb}
The SKEM and PSKEM classifier algorithms were also tested on a number of other benchmark data sets including (i) rice varieties \cite{Cinar2019}, consisting of 3810 samples of 7-D features in 2 classes; (ii) ionosphere \cite{Wing2003}, containing 351 samples of 34-D data in 2 classes; (iii) fashion MNIST \cite{Xiao2017} - a one-for-one replacement for MNIST with 60,000 training and 10,000 greyscale test images of 28 x 28 pixels in 10 classes; (iv) CIFAR-10 \cite{Krizhevsky}, which contains 50,000 training and 10,000 test RGB images of 32 x 32 pixels in 10 classes. The rice and ionosphere datasets are available from the UCI archive \cite{UCI}.

{\bf Rice varieties}: With only 7-D features there is no need to partition the data, so only regular SKEM was used for training using 10-fold cross validation (without randomisation). In each trial, a single run of 10 passes was applied to each fold of 90\% with testing on the remaining 10\% of the data. The model with the maximum test accuracy during the 10 passes was retained and the results averaged over the 10 folds. Each run used random initial mean values uniform on [$-1$, 1] with standard deviation $\sigma=2$. The number of components $K$ was varied from 4 to 20, with values around 12 - 14 yielding the best accuracies. Specifically $K=14$ gave 95.0\%$\pm$0.39\% (1-sigma) accuracy assessed on 50 trials. Comparison with conventional techniques was effected using the Matlab Machine Learning toolbox (MMLT). Logistic regression, linear discriminant analysis (LDA) and SVM (both linear and RBF) all gave 93.0\% accuracy, in line with testing in \cite{Cinar2019}.
PSKEM thus outperformed conventional approaches by around 2\% on this data set.

{\bf Ionosphere}: Extracting features 3 to 34 gives a 32-D data set. PKEM partitions were allocated as 1x32, 2x16, 4x8, 8x4 and 16x2 dimensional. To allow a direct comparison with \cite{Titsias}, training was carried out with randomised 5-fold cross validation. In the latter, a special case PRBF similar to MDA with a spherical covariance for each mixture component $P_k=\sigma_k^2 I$ attained an accuracy of 90.6\% with $K=12$ components. Each PSKEM trial used 40 passes with uniform initial mean on [$-1$, 1] and standard deviation $\sigma=10^5$ for each 12-component mixture. The best partition was 2x16-D, which achieved an accuracy of 98.0$\pm$1.2\% (1-sigma) accuracy on 200 trials. The top three approaches using the MMLT were RBF SVM (kernel scale 5.7) 93.7\%, bagged trees 92.9\% and quadratic SVM 92.6\%. PSKEM performance on this data set was therefore around 4\% better than conventional approaches and significantly better than the EM algorithm variants in \cite{Titsias}.

{\bf Fashion MNIST}: PCA features were extracted from the vectorised training image set. Various configurations of PSKEM were implemented having between 3 and 20 blocks of 10- to 20-dimensional features with a maximum product of $mR=200$. The number of components $K$ was 60, 80 or 100. Initial mixture means were uniform on [$-2$, 2] with standard deviation $\sigma=2$. The best PSKEM accuracy obtained was 85.12\% using 60,000 training samples, $K=100$, and 10x15-D blocks, with typical accuracy around 84.5\%. To ensure a fair comparison, benchmarking was carried out using the same 150 PCA features via the MMLT. The following accuracies were obtained: 85.97\% (100 bagged trees); 85.80\% (50 bagged trees); 85.32\% (linear SVM); 84.79\% ($k=5$ nearest neighbour); 84.63\% (kNN $k=1$); 84.33\% (kNN $k=9$); 80.67\% (LDA). Thus PSKEM had similar performance to kNN but was inferior to bagged trees and SVM on this data set. In the absence of a parallel processor implementation, the high computational load for SKEM training on this data set favours conventional classification methods.

{\bf CIFAR-10}: PCA features were extracted from the stacked RGB channels. We obtained 46.02\% accuracy using a PSKEM with $R=4$ blocks of $m=24$ features (96 in all) and $K=80$ components trained on all 50,000 images. Using more than 4 blocks did not improve performance (up to 200 features). It is known that the best accuracy for conventional machine learning approaches using PCA components is around 50\% on CIFAR-10 using SVM classifiers \cite{Abouelnaga}. Deep convnets can achieve $\geq 95\%$ accuracy, exhibiting significant shift and scale invariance in the construction of their feature maps. In contrast, regular machine learning techniques, including SKEM, are sensitive to object size, position and orientation in the image, which is not standardised in CIFAR-10.

{\bf Summary}: Table \ref{Tab1} contains a summary of the comparative simulation results on the 5 data sets. Column 2 details the main PSKEM algorithm parameters used to obtain the accuracy (Acc1) in column 4. The feature dimension can be obtained as the product $mR$. The CPU time for training is indicated in coumn 3 in seconds per pass of the EM algorithm, as obtained on an Intel Core i5 3.3 GHz CPU with 8 GB RAM. Column 5 (Acc2) is the accuracy of the best conventional classifier method tested in cases where the MMLT was used. Here, ``conventional'' means any approach not using convolutional neural networks. A reference is provided where the result is external to this study. Column 6 details the best method of those known or tested (B-trees means bagged trees).

\begin{table}
\center
\begin{tabular}{|c|c|c|c|c|c|}
\hline
Data set & $R\times m,K$ & Time & Acc1 \% & Acc2 \% & Best \\
 &  & (s)/pass & & & method \\
\hline
Rice & 1x7, 14 & 0.25 & 95.0 & 95.0 & SKEM\\
\hline
Iono & 2x16, 12 & 0.083 & 98.0 & 98.0 & PSKEM\\
\hline
MNIST & 3x13, 60 & 26 & 97.48 & 99.48 & k-NN \cite{Decoste2002}\\
\hline
F-MNIST & 10x15, 100 & 238 & 85.12 & 85.97 & B-trees\\
\hline
CIFAR-10 & 4x24, 80 & 274 & 46.0 & 50.0 & SVM \cite{Abouelnaga} \\
\hline
\end{tabular}
\caption{Performance of PSKEM versus conventional classifier methods. Key: Acc1 - PSKEM accuracy; Acc2 - best achieved.}
\label{Tab1}
\end{table}

\subsection{Trade-off Studies}\label{tos}
Further experiments were undertaken to better understand the trade-offs involved in the implementation and tuning of the PSKEM algorithm and its performance relative to the regular (unpartitioned) SKEM. These simulations were carried out on half-size (14 x 14 pixel) MNIST images for simplicity, using 30,000 training images, 10,000 test images and 36 PCA features. The following trade-offs were investigated: (i) mean accuracy as a function of block size $m$ (SKEM feature dimension) and number of blocks $R$ for varying number of components $K$, with the product $mR$ fixed at 36 and blocks obtained by sequential partitioning; (ii) as in (i) but for interleaved partitions and (iii) for random (non-overlapping) partitions. For instance with 3 blocks, interleaving gives partitions of the form [1,4,7,...], [2,5,8,...], [3,6,9,...] in terms of feature indexes.

\begin{figure}
\center
\includegraphics[height=8cm]{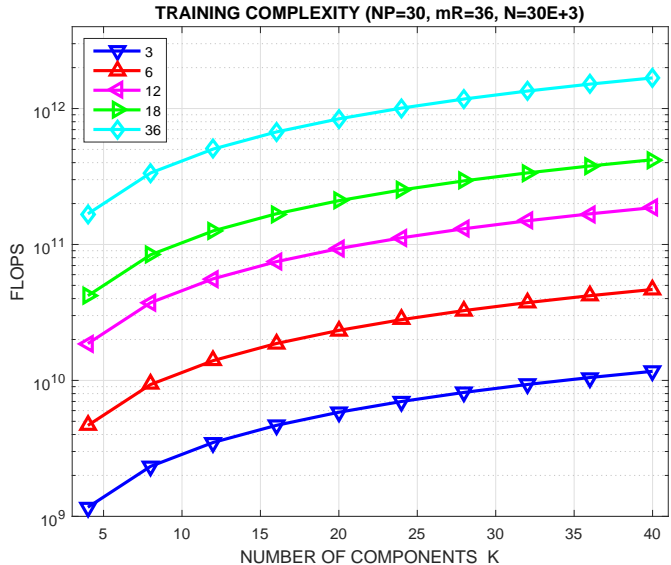}
\caption{Training complexity versus $K$ and partition dimension $m$ for 36-D features, 30 passes \& 30,000 samples. Unpartitioned SKEM is $m=36$.}
\label{fig1}
\end{figure}

In each case, 20 runs with different initial mixture means were performed to estimate the average accuracy (a single point on one of the plots). The choice of 36 features allowed the following partitions to be formed: 12 x 3-D, 6 x 6-D, 3 x 12-D, 2 x 18-D and 1 x 36-D. Notice that the 36-D PSKEM with one partition is in fact an unpartitioned SKEM. To give some idea of the computational complexity, the approximate operation count for NP=30 passes of the different partitioned SKEMs is shown in Fig. \ref{fig1} as a function of $K$. Note that these figures account for the entire training regime. The 36-D SKEM required more passes to give peak performance, so NP=40 was used in the latter case. In practice, only the regular SKEM is not amenable to parallel processing.

\begin{figure}
\center
\includegraphics[height=8cm]{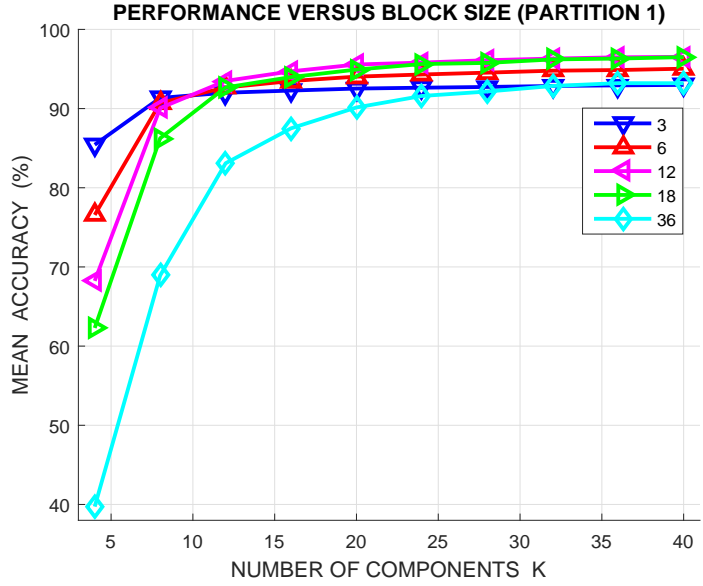}
\caption{Mean accuracy versus $K$ and partition dimension $m$ (sequential partitions). Unpartitioned SKEM is $m=36$.}
\label{fig2}
\end{figure}

\begin{figure}
\center
\includegraphics[height=8cm]{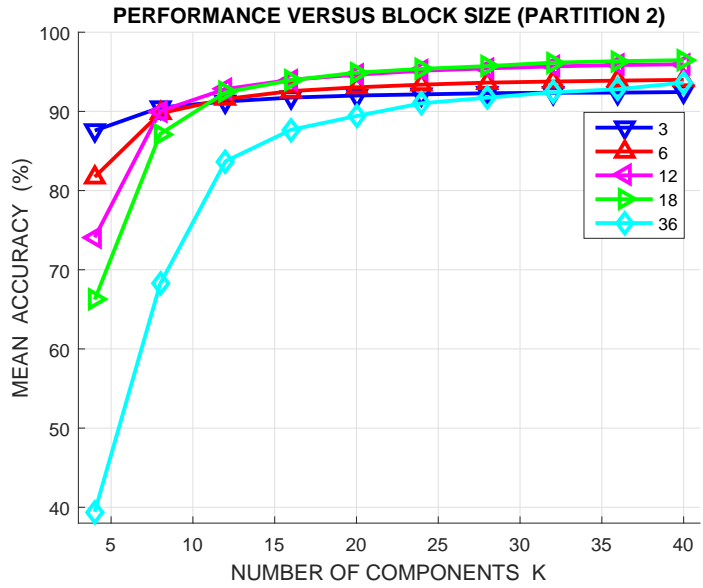}
\caption{Mean accuracy versus $K$ and partition dimension $m$ (interleaved).}
\label{fig3}
\end{figure}

\begin{figure}
\center
\includegraphics[height=8cm]{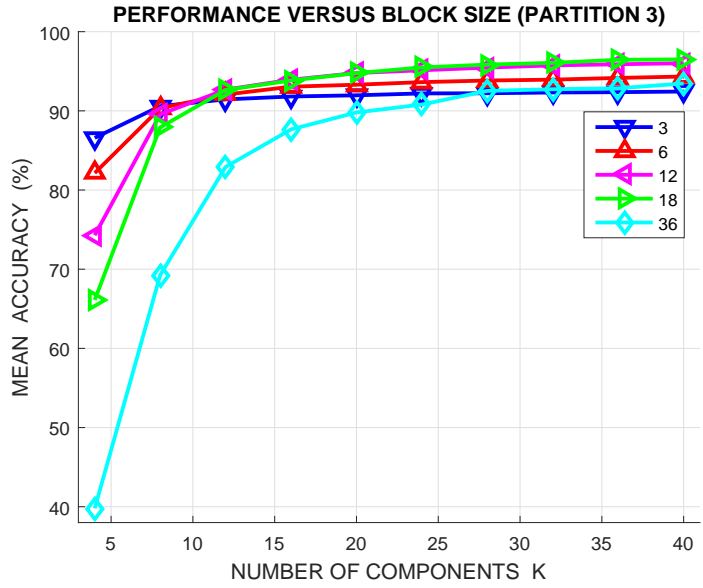}
\caption{Mean accuracy versus $K$ and partition dimension $m$ (random).}
\label{fig4}
\end{figure}

Figs. \ref{fig2} - \ref{fig4} show the mean accuracy results for the 3 different partitioning arrangements (1) sequential, (2) interleaved and (3) random. Standard deviation (omitted from plots for clarity) across the 5 partitions ($m=3$ to $m=36$) was approximately (0.6, 2, 3, 3, 0.6)\% at $K=4$ and (0.1, 0.1, 0.1, 0.3, 0.8)\% at $K=40$. The only significant differences between these groups of curves is for small number of components $K$ (e.g. the 12 x 3-D PSKEM), which seems to favour interleaved and random partitions over sequential.
No numerical divergences (where a covariance matrix becomes singular) were observed for any of the random initial configurations. Empirically, the numerical stability of the SKEM algorithm appears to depend more on the number of components than the initial choice of mixture means.

Concerning relative performance, which is monotonically increasing in $K$, for larger $K$, performance levels off at around 96\% for all partitioning arrangements. The best performance was obtained in all cases for partitions with small $R$ and large $m$, i.e., 3 x 12-D and 2 x 18-D. The performance of the unpartitioned SKEM, whose computational requirements are more than 5 times greater than all other PSKEMs tested, was worse than even the 12 x 3-D PSKEM except at large $K\geq 28$. This convincingly establishes the superiority of the PSKEM over the regular SKEM both in terms of performance and complexity.

Further evidence of the efficiency of the PSKEM algorithm is furnished by comparing it with the MDA equivalent, which assumes an identical covariance for all mixture components and classes. The MDA update equations given by (\ref{skemw})--(\ref{skemmu}) and (\ref{mdacov}) were also implemented. To ensure a fair comparison, both the PSKEM and MDA used the same initial mixture, the means of which were varied on each run. On the half-size MNIST data, using 3 x 13-D partitions with 30,000 samples and 30 passes, the following $\mu\pm 3\sigma$ accuracies were obtained on 100 runs with $K=20$ components: PSKEM $0.9552\pm 0.0143$; MDA $0.9154\pm 0.0093$; and $0.9662\pm 0.0053$ (PSKEM) and $0.9340\pm 0.0065$ (MDA) with $K=40$.

\begin{figure}
\center
\includegraphics[height=8cm]{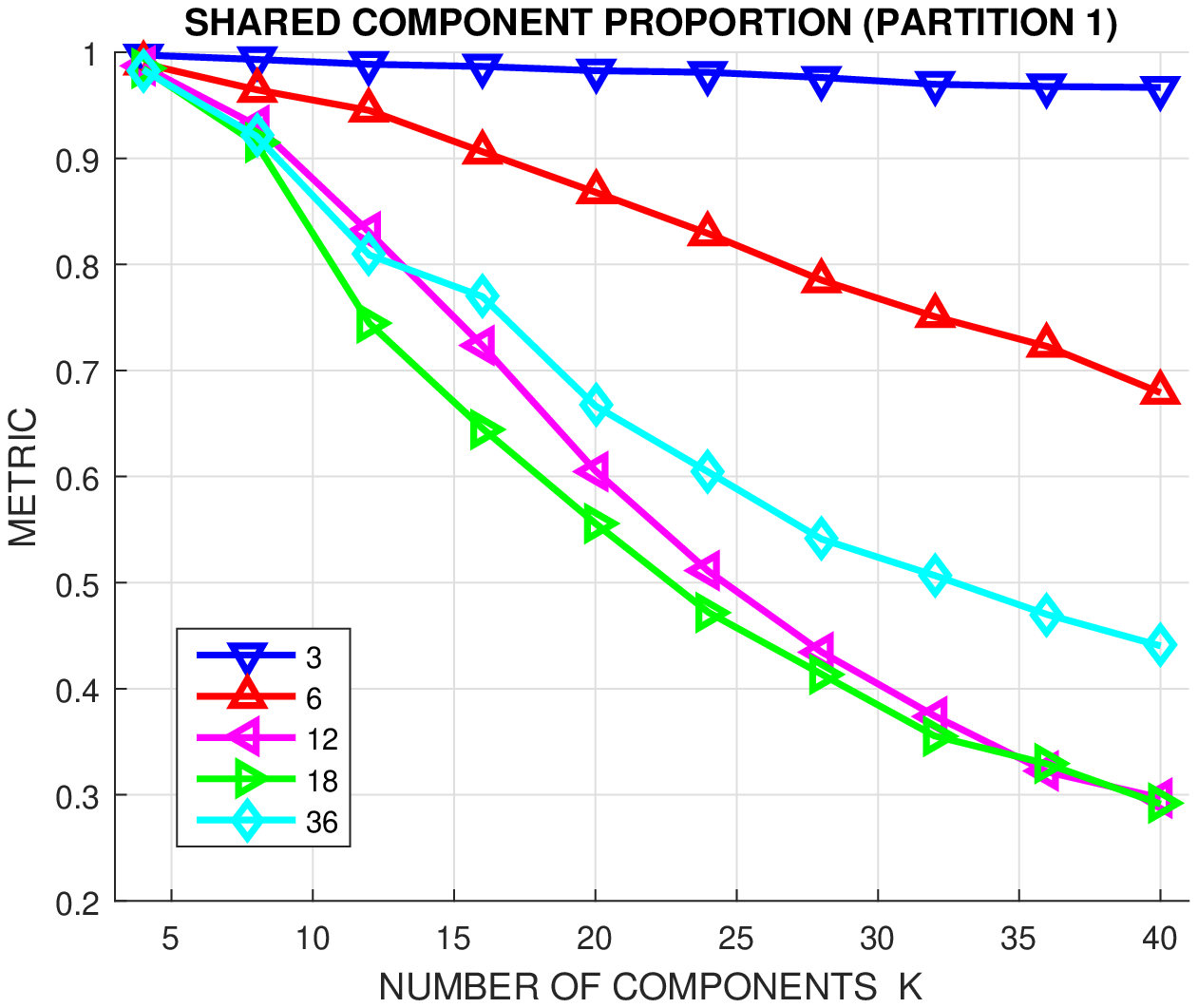}
\caption{Mean proportion of shared mixture components versus $K$ and $m$.}
\label{fig5}
\end{figure}

The trade-off testing produced a large quantity of converged SKEM mixtures, enabling further analysis. Although direct visualisation of the Gaussian mixtures is not practical except in low dimensional cases (see \cite{Pulford2020a} for examples), it is easy to examine the mixture weights $\pi_{ki}$ (\ref{skempi}) to ascertain the extent of component sharing and see if this relates to mean accuracy performance. Specifically, we state that two classes $i$ and $j$ share a mixture component $k$ if both $\pi_{ki}\geq\tau$ and $\pi_{kj}\geq\tau$ where $\tau$ is a significance threshold taken as 0.01. Given a matrix of mixture weights $\Pi=(\pi_{ki})$, $i=1,\ldots,L$, $k=1,\ldots,K$, we form $A=(a_{ki})$ by thresholding, i.e., $a_{ki}=1$ if $\pi_{ki}\geq\tau$ and 0 otherwise. The row sums of $A$ are $a_k=\sum_{i=1}^{L}a_{ki}$. Seeking a metric for the proportion of shared components, we define two positive integers $N_T=\sum_{k=1}^{K}a_{k}$ and $N_S=\sum_{k=1}^{K}(a_{k}|\,a_{k}>1)$. The {\em shared proportion} is then $N_S/N_T$, which is between 0 and 1. For multiple blocks we sum both $N_S$ and $N_T$ before taking the ratio; while for multiple runs, we average the metrics obtained on each run. The result of this exercise is shown for the PSKEM with sequential partitioning in Fig. \ref{fig5}. As expected, all PSKEM variants tend to share fewer mixture components as $K$ increases. However, there are also some surprises. The two variants 3 x 12-D and 2 x 18-D, which have the best performance for $K>12$, also share components the least: for large $K$, there is around 50\% less component sharing on average than the unpartitioned SKEM, which may be limiting the performance of the latter.

\section{Conclusions and Further Work}\label{conc}

We have considered the application of the EM algorithm for supervised training of shared kernel Gaussian mixture models. Despite the solid theoretical foundations and advantages of the EM algorithm, there is little evidence of its application in the last 10 years for neural network training. It seems that there are both theoretical and practical reasons for the abandonment of this approach. We provided a rigorous proof of the shared kernel EM algorithm based on data association theory, which uses categorical latent variables. This should help researchers wishing to extend EM-based techniques, for instance, to situations where the hidden variables are dependent. The use of class-conditioned indexing also simplifies the algorithm statement compared with previous works.

We developed a reduced complexity partitioned SKEM algorithm. With $R$ partitions of feature variables, PSKEM is $R^2$ times more efficient than non-partitioned SKEM. When the partitions are statistically independent, this is an exact representation of the joint data likelihood. The partitioned SKEM can be applied to much larger data sets than earlier supervised EM algorithms. We studied the performance of PSKEM on the MNIST data set as a function of the block size and number of blocks, the number of components per kernel, and the partitioning arrangement. The results indicate the that the PSKEM, in addition to significantly lower computational requirements, is numerically robust and has consistently better classification accuracy than the non-partitioned SKEM. 

Good performance was obtained on MNIST. On two lower dimensional data sets (rice and ionosphere) the SKEM outperformed conventional (non-convnet) classification methods. Accuracy on fashion MNIST (for the same feature set) was similar to k-nearest neighbours and slighly below bagged decision trees and SVM. Performance on CIFAR-10 was around 5\% lower than that attainable with SVM, suggesting that SKEM is more sensitive to object size and pose.

A natural direction for future research is the initialisation method. For instance, K-means clustering or standard EM could be used to ``warm start'' the SKEM algorithm (see, for example, \cite{Raitoharju}). The k-d tree algorithm in \cite{Hoori} is another candidate initialisation approach. An important question is the degree to which the performance of the partitioned SKEM depends on the independence of the feature partitions. A starting point for further investigation is the application of independent component analysis (ICA) \cite{Dhir}. Since the partitioned SKEM is inherently parallelizable, multi-processor implementations are also possible, which would alleviate the computational bottleneck for training on larger data sets.

SKMs, being based on Gaussian mixtures, are good candidates for performance analysis. The kernels are based on soft clusterings of the input feature data and so are directly related to the data set. It should be possible to obtain, at least approximately, analytical expressions for confusion matrix error probabilities and decision hypersurfaces, which bodes well for obtaining ``explainability'' results for SKMs. This contrasts with ``black box'' machine learning approaches like SVMs and convolutional neural networks, whose classification decisions are typically hard to interpret \cite{Townsend}.

\bibliographystyle{unsrt}
\bibliography{SKEM_refs}

\section*{Appendix A: Binary indicators versus categorical hidden variables}
We examine the construction of the auxiliary function $Q(\Theta,\Theta_0)$ for the standard EM algorithm defined in (\ref{baum2a}) using binary indicator vectors, as in \cite{Titterington}, and via scalar categorical data association variables, as used in this paper, to highlight their differences. $K$-D binary vector indicators $\bfz_i$ have components $z_{ij}=1$ when data point $x_i$ is from mixture component $j$, and $z_{ij}=0$ otherwise. As in section \ref{skemdetails}, categorical variables $z_i=k$, which are scalar, associate component $k$ to $x_i$. While the difference may appear trivial, only the latter leads to manageable expressions when standard probability theory rules are applied. The reason, as we will see, is that for a given $\bfz_i$, only one component can equal 1 at a time, which makes all the components dependent random variables.

For the binary vector case, we have from (\ref{baum2a}):
\[
Q_B(\Theta,\Theta_0)=\sum_{\bfz_1}\cdots\sum_{\bfz_N}\Pr(Z|X,\Theta_0)\log{\rm p}(X,Z|\Theta)
\]
where a sum over a vector index means
\[
\sum_{\bfz_i}=\sum_{z_{i1}=0}^{1}\cdots\sum_{z_{iK}=0}^{1}
\]
subject to the ``one-at-a-time'' constraint mentioned before. Thus $Q_B(\Theta,\Theta_0)$ contains $2^{KN}$ terms, only $K^N$ of which are non-zero.
Owing to the independence of the indicators for different data points (which are assumed {\em iid}):
\begin{eqnarray*}
\log{\rm p}(X,Z|\Theta)&=&\sum_{i=1}^{N}\log{\rm p}(x_i,\bfz_i|\Theta)\\
\Pr(Z|X,\Theta_0)&=&\prod_{j=1}^{N}\Pr(\bfz_j|x_j,\Theta_0)
\end{eqnarray*}
Now the components of each $\bfz_i$ are not independent so we cannot further factorise the preceding expressions to simplify them. We also note that ``power law'' form in (\ref{titlik}) does not arise naturally from the standard probability definitions applied here. However, without actually starting from (\ref{titlik}) (which is the usual route for EM derivations), $Q_B(\Theta,\Theta_0)$ is unwieldly and difficult to simplify. This provides insight into why many conventional treatments on EM seem to skip to a direct expression of the $Q(\Theta,\Theta_0)$ function.

For categorical data association variables $Z=(z_1,\ldots,z_N)$, where the (scalar) components $z_i$ are independent, we have:
\begin{eqnarray*}
\log{\rm p}(X,Z|\Theta)&=&\sum_{i=1}^{N}\log{\rm p}(x_i,z_i|\Theta)\\
\Pr(Z|X,\Theta_0)&=&\prod_{j=1}^{N}\Pr(z_j|x_j,\Theta_0)
\end{eqnarray*}

Defining $g(z_i,x_j)=\log{\rm p}(z_i,x_j|\Theta)$ and $h(z_i|x_j)=\Pr(z_i|x_j,\Theta_0)$, the definition (\ref{baum2a}) gives:
\[
Q_C(\Theta,\Theta_0)=\sum_{z_1=1}^{K}\cdots\sum_{z_N=1}^{K}\sum_{i=1}^{N}g(z_i,x_i)\prod_{j=1}^{N}h(z_j|x_j)
\]
After some straightforward manipulations contained in \cite{Pulford2020b} (or \cite{Bilmes} with different notation), which are analogous to the derivations in (\ref{Lsums})--(\ref{baum2e}) and  omitted for brevity, there results
\[
Q_C(\Theta,\Theta_0)=\sum_{n=1}^{N}\sum_{k=1}^{K}\Pr(z=k|x_n,\Theta_0)\log{\rm p}(z=k,x_n|\Theta)
\]
Thus it can be appreciated that the use of binary indicator vectors results in dependent data association events due to redundancies in the hidden variables. These dependencies are not simple to resolve without actually expressing the likelihood in terms of powers of PDFs as in (\ref{titlik}) rather than the much simpler expression in (\ref{pycat}) obtained via standard data association. Admittedly, without going to this level of detail, it is not obvious that such issues are present. On the other hand, categorical hidden variables do not require any special probabilistic considerations and are theoretically rigorous for the evaluation of the $Q$ function both for SKEM and in other settings, e.g. \cite{Pulford2002}.

\section*{Appendix B: Structure of the Partitioned SKM}
To clarify the construction of the partitioned SKM in equation (\ref{pskemfact}), we consider a simple example with $R=2$ partitions with $K=2$ components each. By assumption, each partitioned density is a GMM of the form
\[
q_{rj}(x_i)=\sum_{k=1}^{K}\pi_{kj}^{(r)}\,{\calN}\{x_r;\mu_k^{(r)},P_k^{(r)}\},~r=1,2,~j=1,\ldots,L
\]
where $m_1={\rm dim}(x_1)$ and $m_2={\rm dim}(x_2)$ satisfy $m_1+m_2=M$ where $M={\rm dim}(\bfx)$.
By assumption, the joint class-conditioned PDF is given by
\[
{\rm p}(\bfx|c=j)=q_{1j}(x_1)\,q_{2j}(x_2)
\]
When $x_1$ and $x_2$ are independent Gaussian random variables:
\[
{\calN}\{x_1;\mu_1,P_1\}{\calN}\{x_2;\mu_2,P_2\}={\calN}\{
\left[
\begin{array}{c}
x_1 \\ x_2
 \end{array}
\right]; %[x_1;x_2];
\left[
\begin{array}{c}
\mu_1 \\ \mu_2
\end{array}
\right], %[\mu_1;\mu_2],
P_1\oplus P_2\}
\]
where $P_1\oplus P_2$ is the direct sum of matrices $P_1$ and $P_2$, \ie, the block diagonal matrix formed by the latter.  The product PDF $q_{1j}(x_1)\,q_{2j}(x_2)$ with $K=R=2$ is expressible as:
\begin{eqnarray*}
&&\hspace{-15mm} \sum_{k=1}^{2}\pi_{kj}^{(1)}\,{\calN}\{x_1;\mu_k^{(1)},P_k^{(1)}\}
\sum_{k=1}^{2}\pi_{kj}^{(2)}\,{\calN}\{x_2;\mu_k^{(2)},P_k^{(2)}\}\\
 &=& \pi_{1j}^{(1)}\pi_{1j}^{(2)} 
{\calN}\{\bfx;
\left[
\begin{array}{c}
\mu_1^{(1)} \\ \mu_1^{(2)}
\end{array}
\right],P_1^{(1)}\oplus P_1^{(2)}\} \\
 &&+~ \pi_{1j}^{(1)}\pi_{2j}^{(2)} 
{\calN}\{\bfx;
\left[
\begin{array}{c}
\mu_1^{(1)} \\ \mu_2^{(2)}
\end{array}
\right],P_1^{(1)}\oplus P_2^{(2)}\} \\
 &&+~ \pi_{2j}^{(1)}\pi_{1j}^{(2)} 
{\calN}\{\bfx;
\left[
\begin{array}{c}
\mu_2^{(1)} \\ \mu_1^{(2)}
\end{array}
\right],P_2^{(1)}\oplus P_1^{(2)}\} \\
 &&+~ \pi_{2j}^{(1)}\pi_{2j}^{(2)} 
{\calN}\{\bfx;
\left[
\begin{array}{c}
\mu_2^{(1)} \\ \mu_2^{(2)}
\end{array}
\right],P_2^{(1)}\oplus P_2^{(2)}\}
\end{eqnarray*}
Clearly this has $2^2=4$ Gaussian PDF terms instead of the original 2-term M-dimensional GMM. However, it is also less general since it ignores correlations between the variables in different partitions.

\newpage
\setlength\parindent {0pt}
\setlength\parskip {3pt}

\end{document}